\documentclass[10pt,twocolumn]{article}
\usepackage{simpleConference}
\usepackage{framed,multirow}
\usepackage[sort&compress,square,comma,authoryear]{natbib}
\usepackage{amssymb}
\usepackage{latexsym}

\usepackage{url}
\usepackage{xcolor}
\usepackage{graphicx}

\usepackage{amsmath,amssymb}
\usepackage{algorithm2e}
\usepackage{verbatim}
\usepackage{color,soul}
\usepackage{hyperref}

\DeclareMathOperator*{\argmin}{argmin}

\definecolor{newcolor}{rgb}{.8,.349,.1}

\DeclareMathOperator{\E}{\mathbb{E}}

\begin{document}

\title{End-to-end learning of convolutional neural net and dynamic programming for left ventricle segmentation}

\author{Nhat M. Nguyen\\
\texttt{nmnguyen@ualberta.ca}\\
\\
Computing Science\\
University of Alberta, Canada\\
\and
Nilanjan Ray\\
\texttt{nray1@ualberta.ca}\\
\\
Computing Science\\
University of Alberta, Canada\\
}

\maketitle
\thispagestyle{empty}

\begin{abstract}
Differentiable programming is able to combine different functions or programs in a data processing pipeline with the goal of applying end-to-end learning or optimization. A significant impediment for differentiable programming is the non-differentiable nature of algorithms. We propose to use synthetic gradients (SG) to overcome this difficulty. SG uses the universal function approximation property of neural networks. We apply SG to combine convolutional neural network (CNN) with dynamic programming (DP) in end-to-end learning for segmenting left ventricle from short axis view of heart MRI. Our experiments show that end-to-end combination of CNN and DP requires fewer labeled images to achieve a significantly better segmentation accuracy than using only CNN.
\end{abstract}



\section{Introduction}
Recent progress in medical image analysis is undoubtedly boosted by deep learning \citep{Greenspan, DL_MIA}. Progress is observed in several medical image analysis tasks, such as segmentation \citep{Lesion_seg,Tumor_seg}, registration \citep{GHOSAL201781}, tracking \citep{Cell_tracking} and detection \citep{Deep_detection}.

Deep learning has been most successful where plenty of data was annotated, e.g., diabetic retinopathy \citep{retinopathy}. For many other applications, limited amount of labeled / annotated images pose challenges for deep learning \citep{Greenspan}. Transfer learning is the dominant approach to deal with limited labeled data in medical image analysis, where a deep network is first trained on an unrelated, but large dataset, such as Imagenet; then the trained model is fine-tuned on smaller data set specific to the task. Transfer learning has been applied for lymph node detection and classification \citep{Transfer}, localization of kidney \citep{Ravishankar} and many other tasks \citep{Tajbakhsh}. Data augmentation is also applied to deal with limited labeled data \citep{Unet}.

To overcome the lack of limited labeled data, a complementary approach uses prior knowledge about the segmentation problem \citep{ShapePriorCNN}. However, CNN itself lacks a mechanism to incorporate such prior knowledge. Hence, there is a need to combine CNN with classical segmentation methods, such as active contours and level set methods \citep{Acton2009} so that the latter can directly incorporate adequate prior knowledge.

Toward incorporating traditional segmentation within deep learning, \cite{DeepLevelset} proposed to use CNN to learn a level set function (signed distance transform) for salient object detection. \cite{TangVZCJ17} used level set in conjunction with deep learning to segment liver CT data and left ventricle from MRI. Deep active contours \citep{DeepAC} combined CNN and active contours. \cite{NGO2017159} combines level set and CNN to work with limited labeled data for left ventricle segmentation. However, these works fell short of an end-to-end training process that offers the advantage of not having to deal with a complex training process involving multiple types of annotations.

End-to-end learning, which is not yet abundant in medical image analysis, has been utilized for level set and deep learning-based object detector \citep{LevelsetRNN} that modeled level set computation as a recurrent neural network. \cite{EndToEndAC} have combined CNN and active contours in end-to-end training with a structured loss function. \cite{Ghosh2017} uses principal components analysis along with CNN in end-to-end learning to incorporate object shape prior for segmentation.

\begin{figure*}[t]
\centering
\includegraphics[width=\textwidth]{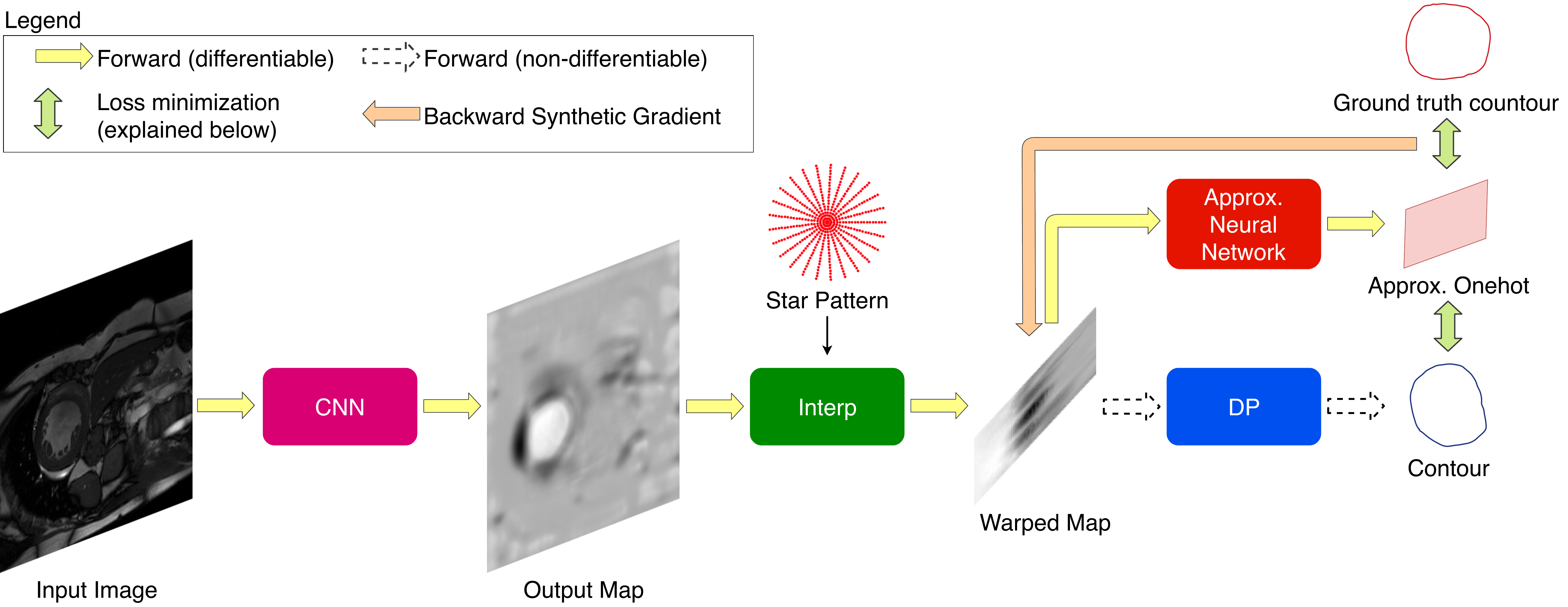}
\caption{Proposed method: EDPCNN.} \label{pipeline}
\end{figure*} 

While end-to-end learning or differentiable programming \citep{Baydin} often provides a better accuracy in various tasks, it comes with a significant limitation - all modules or components need to be differentiable \citep{Glasmachers17}. Most likely, this stringent requirement has limited the use of mixing traditional image analysis algorithms with deep learning in medical image analysis.

In this work, we demonstrate how to combine \textbf{both differentiable and non-differentiable} modules together in an end-to-end learning. Our use case is left ventricle segmentation that combines CNN with active contours. We compute active contours using dynamic programming (DP) \citep{DPSnake}. While CNN is differentiable, DP is non-differentiable in nature due to the presence of \textit{argmin} function in the algorithm. Further, we demonstrate that end-to-end combination of CNN and DP for left ventricle segmentation \textbf{can overcome the lack of annotated data to a significant extent.}

We use a neural network as a bypass for a non-differentiable module. The bypass network approximates the output of the non-differentiable module and its subgradient using the universal function and generalized gradient approximation property \citep{Hornik1990UniversalAO}. Backpropagation uses gradient of the bypass network as a proxy for the subgradient of the non-differentiable module. This technique known as synthetic gradients (SG) has been used before for fast and asynchronous training of differentiable modules \citep{SG2016}. In this work we show that SG can be successfully applied across a non-differentiable DP module.

\begin{figure*}[t]
\centering
\includegraphics[width=\textwidth]{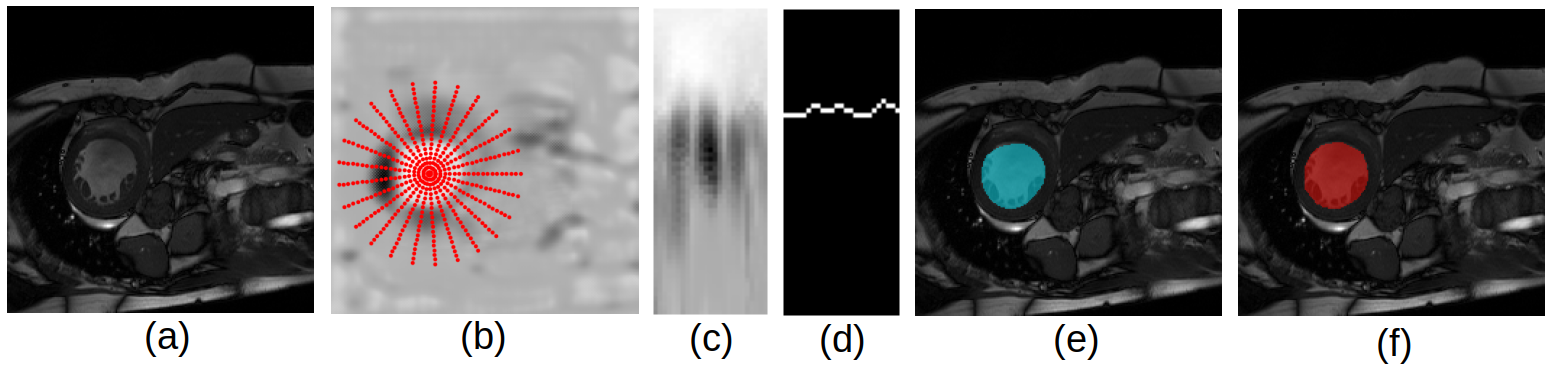}
\caption{Illustrations of processing pipeline: (a) input image, (b) Output Map with an example star pattern, (c) Warped Map and (d) output indices indicating LV on the warped space (e) segmentation obtained with EDPCNN (f) ground truth.} \label{example_result}
\end{figure*}

\section{Proposed Method}

In this paper, we propose end-to-end dynamic programming and convolutional neural networks (EDPCNN) to segment left ventricle from short axis MRI \citep{LV_seg}. Fig. \ref{pipeline} illustrates our processing pipeline. The input to the CNN (we use U-Net \citep{Unet} in our experiments) is an MR image as shown in Fig. \ref{example_result}(a). Output from the CNN is a processed image, called output map, on which a pattern is overlaid in Fig. \ref{example_result}(b). The pattern consists of a few graduated radial lines. We refer to it as a ``star pattern.'' The interpolator (``Interp'' in Fig. \ref{pipeline}) interpolates output map on the points of the star pattern and warp the interpolated values in a matrix called ``Warped Map'' in Fig. \ref{pipeline}. Fig. \ref{example_result}(c) illustrates a Warped Map. DP minimizes a cost function on the Warped Map and chooses exactly one point on each radial line in the star pattern to output a set of indices in the warped domain as shown in Fig. \ref{example_result}(d). Mapping the indices back to the image space gives us a closed contour as the final segmentation, as shown in Fig. \ref{example_result}(e). In comparison, ground truth segmentation, created by an expert, is shown in \ref{example_result}(f). 

All computations within EDPCNN pipeline are differentiable except for the \textit{argmin} function calls inside the DP module that  render the entire pipeline unsuitable for end-to-end learning using automatic differentiation. For example, if there is a differentiable loss function that measures the error between output contour and ground truth contour, we would not be able to train the system end-to-end, because gradient would not reliably flow back across the \textit{argmin} function using the standard mechanisms of automatic differentiation. In the past, soft assignment has been utilized to mitigate the issue of non-differentiability for the \textit{argmin} function \citep{Soft_assn}. Here, we illustrate SG to approximate the subgradient of cost with respect to the Warped Map, so that all the preceding differentiable layers (Interp and CNN) can apply standard backpropagation to learn trainable parameters. Fig. \ref{pipeline} illustrates that an approximating neural network (``Approx. Neural Network'') creates a \textit{differentiable bypass} for the non-differentiable DP module. This second neural network approximates the contour that the DP module outputs. Then a differentiable loss function is applied between the ground truth contour and the output of the approximating neural network, making backpropagation possible with automatic differentiation. This mechanism is known as synthetic gradients, because the gradients of the approximating neural network serves as a proxy for the gradients of the DP module. In the next two subsections, we discuss DP and SG within the setup of left ventricle segmentation. 
 
\subsection{Dynamic Programming}
Use of DP in computer vision is wide ranging, including interactive object segmentation \citep{Felzenszwalb}. Here, we use the DP setup described by \cite{DPSnake} to delineate star-shaped/blob objects that perfectly describe left ventricles in the short axis view. 

Let the star pattern have $N$ radial lines with $M$ points on each line. DP minimizes the following cost:
\begin{equation}
    \min_{v_1,\dots,v_N} \ E(N,v_N,v_1) + \sum_{n=1}^{N-1} E(n,v_n,v_{n+1}),
\end{equation}
where each variable $v_n$ is descrete and $v_n \in \{1,\dots,M\}$. Cost component for the radial line $n$ is $E(n,i,j)$ and it is defined as follows:
\begin{equation}
\begin{aligned}
    E(n,i,j)= 
    \begin{cases}
        \begin{aligned}
            & g(n,i)-g(n,i-1)+ g(n \oplus 1,j)- \\ & g(n \oplus 1,j-1),
            \text{ if } |i-j| \leq \delta,
        \end{aligned} \\ \\
        \begin{aligned}
            \infty, \text{ otherwise} \\
        \end{aligned}
    \end{cases}
\end{aligned}
\label{eq:eng}
\end{equation}
\noindent where $g$ is the Warped Map in the EDPCNN pipeline (Fig. \ref{pipeline}), with $g(n,i)$ representing the value of Warped Map on the $i^{\text{th}}$ point of radial line $n.$ The symbol $\oplus$ denotes a modulo $N$ addition, so that $N \oplus 1 = 1$ and $n \oplus 1=n+1$ for $n<N.$ The discrete variable $v_n \in \{1,\dots,M\}$ represents the index of a point on radial line $n.$ DP selects exactly one point on each radial line to minimize the directional derivatives of $g$ along the radial lines. The collection of indices $\{v(1),\dots,v(N),v(1)\}$ chosen by DP forms a closed contour representing a delineated left ventricle. To maintain the continuity of the closed contour, (\ref{eq:eng}) imposes a constraint to the effect that chosen points on two consecutive radial lines have to be within a distance $\delta.$ In this fashion, DP acts as a blob object boundary detector maximizing edge contrast, while maintaining a continuity constraint. Algorithm \ref{alg:dp}, which implements DP, can be efficiently vectorized to accommodate image batches suitable for running on GPUs.

\begin{algorithm}
\SetAlgoLined
\tcc{Construct value function $U$ and index function $I$}
 \For{$n=1,\dots,N-1$}{
 \For{$i,k=1,\dots,M$}{
   \eIf{$n==1$}{
        $U(1,i,k) = \min_{1 \leq j \leq M} [E(1,i,j)+E(2,j,k)]$ \;
        $I(1,i,k) = \argmin_{1 \leq j \leq M} [E(1,i,j)+E(2,j,k)]$ \;
   }{
        $U(n,i,k) = \min_{1 \leq j \leq M} [U(n-1,i,j)+E(n+1,j,k)]$ \;
        $I(n,i,k) = \argmin_{1 \leq j \leq M} [U(n-1,i,j)+E(n+1,j,k)]$ \;
  }
  }
 }
 \tcc{Backtrack and output $v(1),\dots,v(N)$}
 $v(1)=\argmin_{1 \leq j \leq M} [U(N-1,j,j)] $\;
 $v(N) = I(N-1,v(1),v(1))$\;
 \For{$n=N-1,\dots,2$}{
    $v(n) = I(n-1,v(1),v(n+1))$\;
 }
 \caption{Dynamic programming}
 \label{alg:dp}
\end{algorithm}

\subsection{Synthetic Gradients}

SG uses the universal function and gradient approximation property of neural networks \citep{Hornik1990UniversalAO}. SG can train deep neural networks asynchronously to yield faster training \citep{SG2016}. In order to use SG in the EDPCNN processing pipeline, as before, let us first denote by $g$ the Warped Map, which is input to the DP module. Let $L(p,p_{gt})$ denote a differentiable loss function which evaluates the collection of indices output from the DP module $p=DP(g)=\{v_1,...,v_N\}$ against its ground truth $p_{gt}=\{v^*_1,...,v^*_n\}$, which can be obtained by taking the intersection between the ground truth segmentation mask and the radial lines of the star pattern. Let us also denote by $F$ a neural network, which takes $g$ as input and outputs a \textit{softmax} function to mimic the output of DP. In Fig. \ref{pipeline}, $F$ apperas as ``Approx. Neural Network." Let $\phi$ and $\psi$  denote the trainable parameters of $F$ and U-Net (``CNN'' in Fig. \ref{pipeline}), respectively. 

The inner minimization in the SG algorithm (Algorithm \ref{alg:sg}) trains the approximating neural network $F$, whereas the outer minimization trains U-Net. Both the networks being differentiable are trained by backpropagation using automatic differentiation. The general idea here is to train $F$ to mimic the output indices of the DP module $p$ as closely as possible, then use $\nabla_gL(F(g),p_{gt})$ to approximate $\nabla_gL(p,p_{gt})$, bypassing the non-differentiable $argmin$ steps of DP entirely. Minimizing $L(p,p_{gt})$ then becomes minimizing $L(F(g),p_{gt})$ with this approximation.

The loss function $L$ in this work is chosen to be the cross entropy between the output of $F$ against the one-hot form of $\{v_1,...,v_N\}$ or $\{v^*_1,...,v^*_N\}$. In this case, $F(g)$ comprises of $N$ vectors, each of size $M$, representing the \textit{softmax} output of the classification problem for selecting an index on each radial line.

We have observed that introducing randomness as a way of exploration in the inner loop of Algorithm \ref{alg:sg} by adding $\sigma \varepsilon_{s}$ to $g$ is important for the algorithm to succeed. Here, $\sigma$ is a hyper-parameter and $\varepsilon_{s}$ is a random vector with its individual components sampled independently from zero-mean, unit variance Gaussian, $\mathcal{N}(0;1)$. Instead of minimizing $L(F(g),DP(g))$, we minimize $L(F(g + \sigma \varepsilon_{s}),DP(g + \sigma \varepsilon_{s}))$. In comparison, the use of SG in asynchronous training by \cite{SG2016} did not have to resort to any such exploration mechanism. 

The correctness of the gradient provided by SG depends on how well $F$ fits the DP algorithm around $g$. We hypothesize that without sufficient exploration added, $F$ will overfit to a few points on the surface and lead to improper gradient signal. Hyperparameter $\sigma$ can be set using cross validation, while the number of noise samples $S$ controls trade off between gradient accuracy and training time. We found that $\sigma=1$ and $S=10$ works well for our experiments.

\begin{algorithm}
\SetAlgoLined
\For{$J,p_{gt} \in$ Training \{Image, Ground truth\} batch} {
    \tcc{Compute Warped Map}
    $g = Interp(Unet(J))$\; 
    \tcc{Train approximating neural network}
    Initialize $s$ to $0$\;
    \While{$s < S$}{
        Sample $\varepsilon_{s}$ from $\mathcal{N}(0;1)$\;
        $\min_\phi L(F(g + \sigma \varepsilon_{s}),DP(g + \sigma \varepsilon_{s}))$\;
        $s = s + 1$\;
    }
    \tcc{Train U-Net}
    $\min_\psi \ L(F(g), p_{gt})$\;
    }
\caption{Training EDPCNN using synthetic gradients}
\label{alg:sg}
\end{algorithm}

\section{Results}

\subsection{Dataset and Preprocessing}
We evaluate the performance of EDPCNN against U-Net on a modified ACDC \citep{bernard2018deep} datatset. As the object centers for the test set is not publicly available, we split the original training set into a training set and a validation set according to \cite{LV_seg}. Following the same work, the images are re-sampled to a resolution of 212 $\times$ 212. As the original U-Net model does not use padded convolution, each image in the dataset has to be padded to size $396 \times 396$ at the beginning, so that the final output has the same size as the original image. After these steps, we remove all images that does not have the left ventricle class from the datasets, resulting in a training set of 1436 images and a validation set of 372 images to be used during training. During evaluation step, the original validation set where the images have not been re-sampled is used.

\subsection{Evaluation Metric}
For evaluation of a segmentation against its corresponding ground truth, we use Dice score \citep{LV_seg}, a widely accepted metric for medical image segmentation. EDPCNN requires the star pattern to be available so that the output of U-Net can be interpolated on the star pattern to produce Warped Map. The star pattern is fixed; but its center can be supplied by a user in the interactive segmentation. For all our experiments, the ground truth left ventricle center for an image serves as the center of the star pattern for the same image. While by design EPDCNN outputs a single connected component, U-Net can produce as many components without any control. Thus, to treat the evaluation of U-Net fairly against EDPCNN, in all the experiments we only select the connected component in U-Net that contains the centroid of the of left ventricle object being evaluated (if none of the regions contains the centroid, select the largest connected component). In addition to Dice score, following \cite{LV_seg}, we also include average symmetric surface distance (ASSD) and Hausdorff distance (HD) in our evaluation experiments. While a higher Dice score is desirable, for both ASSD and HD lower numbers indicate better results.

\subsection{Training Details and Hyperparameters}

We train U-Net and EDPCNN using Adam optimizer \citep{kingma2014adam} with $\beta_1=0.9$, $\beta_2=0.999$, and a learning rate value of 0.0001 to make the training of U-Net stable. Training batch size is 10 for each iteration and the total number of iteration is 20000. No learning rate decay as well as weight decay are used because we have not found these helpful. We evaluate each method on the validation set after every 50 iterations and select the model with the highest validation Dice score.

We use nearest neighbor method to interpolate the output of U-Net on the star pattern to compute Warped Map $g.$ We choose the center of the star pattern for each image to be the center of mass. To make the model more robust and have better generalization, during training, we randomly jitter the center of the star pattern inside the object. We find that this kind of jittering can improve the dice score on smaller training sets by up to about 2\%. We also randomly rotate the star pattern as an additional random exploration. 

The radius of the star pattern is chosen to be 65 so that all objects in the training set can be covered by the pattern after taking into account the random placement of the center during training. The number of points on a radial line has also been chosen to be the radius of the star pattern: $M=65$. For the number of radial lines $N$ and the smoothness parameter $\delta$, we run a grid search over $N \in \{12, 25, 50, 100\}$, $\delta \in \{1, 2, 5, 7, 10\}$ and find $N=50$, $\delta=2$ to be good values. We also find that the performance of our algorithm is quite robust to the choices of these hyperparameters. The Dice score only drops around 3\% when the values of $N$ and $\delta$ are extreme (e.g. $N=100$, $\delta=10$). Lastly, for the optimization of $\min_\phi L(F(g + \sigma \varepsilon_{s}),DP(g + \sigma \varepsilon_{s}))$ in Algorithm \ref{alg:sg}, to make $F(g)$ fit $DP(g)$ well enough, we do the minimization step repeatedly for 10 times.

\begin{figure*}[h]
\centering
\includegraphics[width=\textwidth]{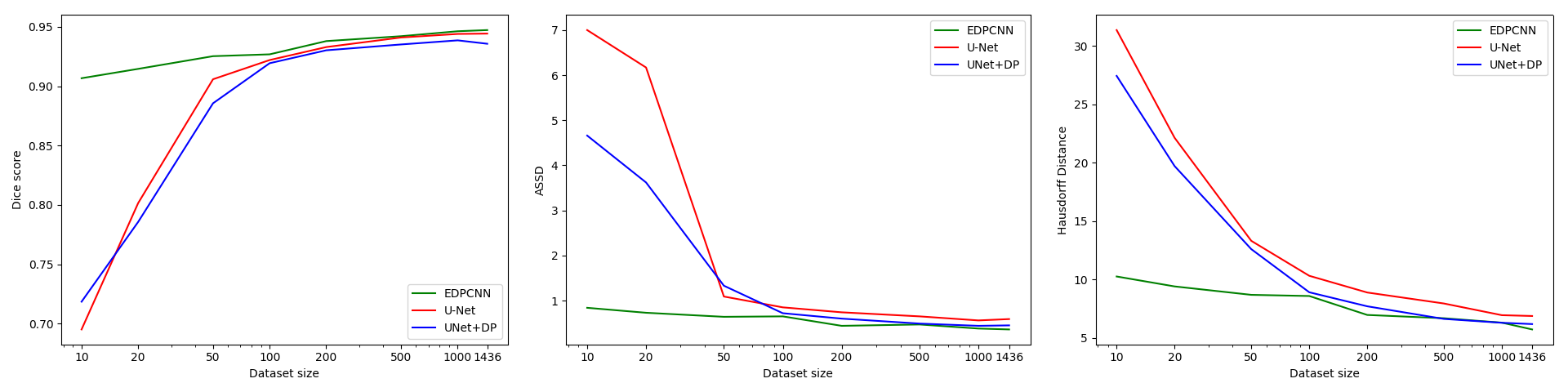}
\caption{Training set size \textit{vs.} Dice, ASSD and Hausdorff distance on validation set.} \label{fig:result_all}
\end{figure*}

The architecture of $F$ used to approximate the output of DP is a U-Net-like architecture. As the size of $g$ is smaller and the complexity of $g$ is likely to be less than the original image, instead of having 4 encoder and 4 decoder blocks as in U-Net, $F$ only has 3 encoder and 3 decoder blocks. Additionally, we use padding for convolutions/transposed convolutions in the encoder/decoder blocks so that those layers keep the size of the feature maps unchanged instead of doing a large padding at the beginning like in U-Net. This is purely for convenience. Note that these choices can be arbitrary as long as $F$ can fit the surface of DP well enough. For the same reason, we find that the number of output channels in the first convolution of $F$, called $base\_channels$, is an important hyperparameter because this value controls the capacity of $F$ and affects how well $F$ fits the surface of DP. We find that $base\_channels=8$ works well for our algorithm (compared to 64 in U-Net). Our code is on \href{https://github.com/minhnhat93/EDPCNN}{Github}.

\subsection{Postprocessing}
As the output contour of DP may sometimes be jagged, we employ a postprocessing step where the output indices are smoothed by a fixed 1D moving average convolution filter with circular padding. The size of the convolutional filter is set using a heuristic to be five. This post-processing also has the effects of pushing the contour to be closer to a circle, which is also a good prior for the left ventricle. This step improves our validation accuracy by around 1.0 percent. Since SG mimics the post-processed output, postprocessing is a part of the end-to-end processing.

\begin{table*}[h]
\centering
\caption{Detailed results for different methods at 10 training samples and full dataset size. Inside bracket is the standard deviation.}
\label{tab:result}
\begin{tabular}{|l|ccc|ccc|}
\hline
        & \multicolumn{3}{c|}{10 training samples}                                 & \multicolumn{3}{c|}{Full dataset}                                       \\ \hline
        & Dice $\uparrow$               & ASSD       $\downarrow$           & HD   $\downarrow$      & Dice     $\uparrow$               & ASSD $\downarrow$  & HD $\downarrow$   \\ \hline
UNet    & 0.695 (0.250)          & 7.66 (11.91)         & 31.34 (23.78)         & 0.944 (0.049)          & 0.59 (0.78)          & 6.88 (4.20)           \\
UNet+DP & 0.719 (0.186)          & 4.66 (4.15)          & 27.42 (17.79)         & 0.936 (0.029)          & 0.45 (0.17)          & 6.19 (2.87)          \\
EDPCNN  & \textbf{0.907 (0.062)} & \textbf{0.84 (0.87)} & \textbf{10.26 (5.36)} & \textbf{0.947 (0.025)} & \textbf{0.36 (0.17)} & \textbf{5.73 (2.89)} \\ \hline
\end{tabular}
\end{table*}

\begin{table*}[h]
\centering
\caption{Detailed results of EDPCNN and UNet for different cardiac phases on validation set when trained with full training set.  Inside parenthesis is the standard deviation}
\label{tab:result_more}
\begin{tabular}{|c|ccc|ccc|}
\hline
& ED (Dice)  & ED (ASSD)  & ED (HD)  &  ES (Dice)  & ES (ASSD)  & ES (HD)  \\ \hline
UNet  & 0.962\,(0.019)  & 0.42\,(0.59)  & 6.34\,(4.79)  & 0.926\,(0.062)  & 0.76\,(0.90)  & 7.42\,(3.42) \\
EDPCNN  & 0.961\,(0.010)  & 0.30\,(0.08)  & 4.84\,(1.56)  & 0.934\,(0.031)  & 0.42\,(0.20)  & 6.62\,(3.56)  \\ \hline
\end{tabular}
\end{table*}

\begin{table*}[h]
\centering
\caption{Computation time on an NVIDIA GTX 1080 TI} \label{tab:time}
\begin{tabular}{|c|c|c|c|c|}
\hline
Method &  Time per tiraining iteration & Total iterations & Total training time & Inference time per image \\
\hline
U-Net  & 0.96s & 20000 & 5h 20m & 0.01465s \\
EDPCNN & 1.575s & 20000 & 8h 45m & 0.01701s \\
\hline
\end{tabular}
\end{table*}

\subsection{Experiments and Discussions}

We train U-Net and EDPCNN increasing training sample size from 10 training images to the full training set size, 1436. To avoid ordering bias, we randomly shuffle the entire training set once, then choose training images from the beginning of the shuffled set, so that each smaller training set is successively contained in the bigger sets, creating telescopic training sets, suitable for an ablation study that is shown in Fig.~\ref{fig:result_all}. 

The ablation experiment (Fig.~\ref{fig:result_all}) demonstrates the effectiveness of combining CNN and DP in an end-to-end learning pipeline. The horizontal axis shows the number of training images and the vertical axis shows the Dice score,  ASSD and HD of LV segmentation on a fixed validation set of images. Note that when the number of training images is small, EDPCNN performs significantly better than U-Net. Eventually, as the training set grows, the gap between the Dice scores, ASSD and Hausdorff distances by U-Net and EDPCNN starts to close. However, we observe that EDPCNN throughout maintains its superior performance over U-Net. 

\begin{figure}[b]
\centering
\includegraphics[width=0.45\textwidth]{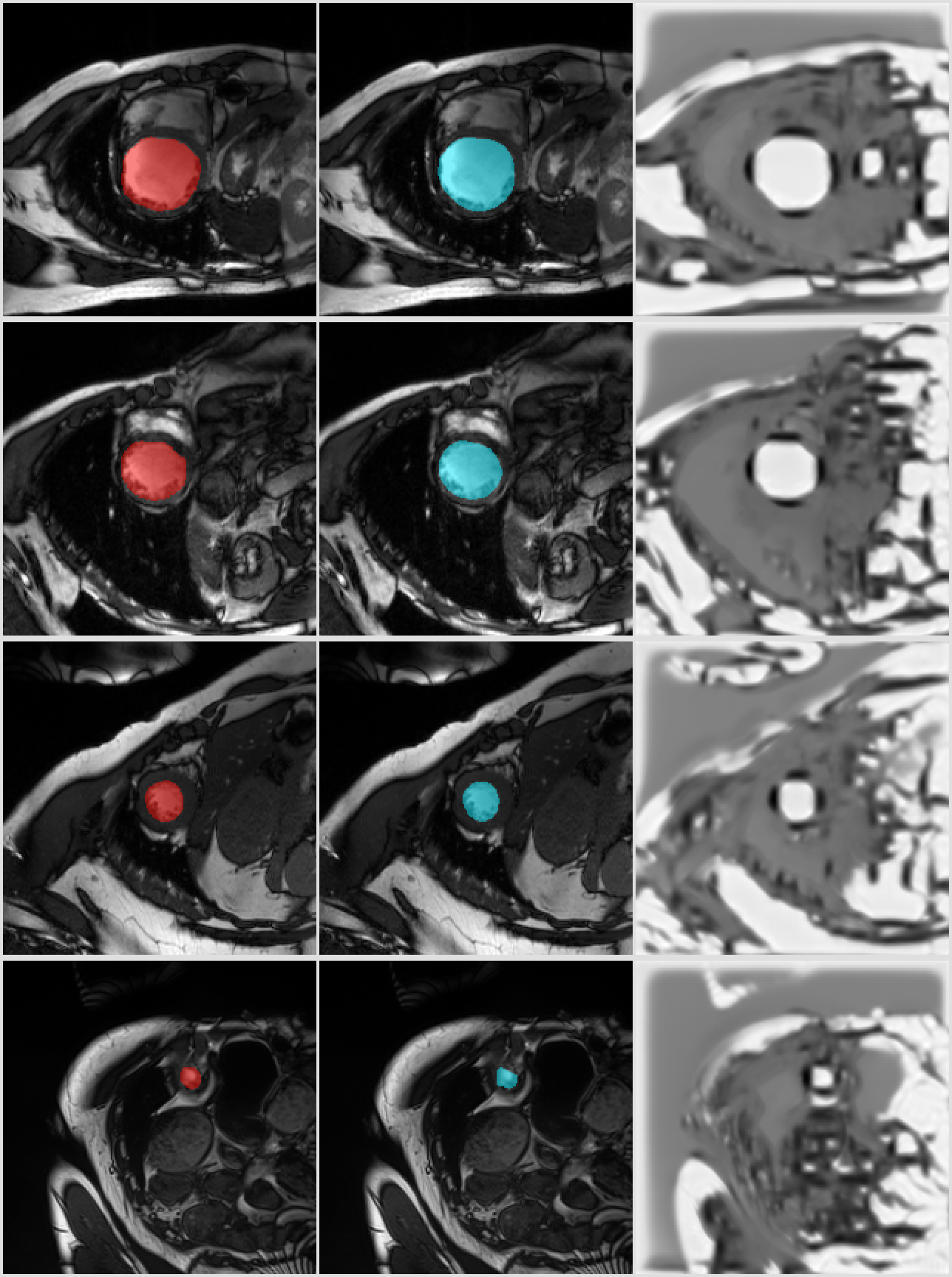}
\caption{Example segmentation. Left to right: ground truth (red), segmentation by EDPCNN (teal), output map of EDPCNN. Top to bottom: object size from large to small.} \label{fig:example_segmentation}
\end{figure}

\begin{figure}[t]
\centering
\includegraphics[width=0.5\textwidth]{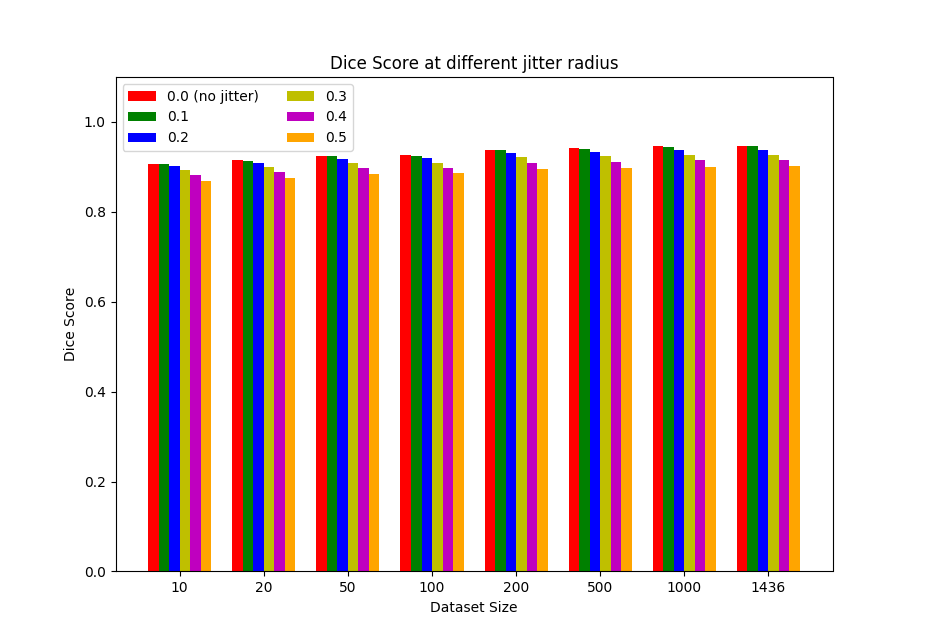}
\caption{Robustness test: Mean Dice score across five runs at different jitter radius.} \label{fig:result_robustness}
\end{figure}

Fig.~\ref{fig:result_all} shows another experiment called ``U-Net+DP''. In the U-Net+DP processing pipeline, DP is applied on the output of a trained U-Net without end-to-end training. Once again, EDPCNN shows significantly better performance than U-Net+DP for small training sets, demonstrating the effectiveness of the end-to-end learning. We hypothesize that DP infuses strong prior knowledge in the training of U-Net within EDPCNN and this prior knowledge acts as a regularizer to overcome some of challenges associated with small training data.

Results on the smallest and full dataset size are shown in Table~\ref{tab:result} that further affirms the sample efficiency of EDPCNN. Note that EDPCNN obtained significantly better results for ASSD and HD in particular, because it is a not pixel-based classification method, such as U-Net. We notice standard deviation is the smallest for EDPCNN. More detalied results for end-diastolic (ED) and end-systolic (ES) cycle are shown in Table \ref{tab:result_more} for training with the full dataset. It can be seen that EDPCNN yielded lower standard deviations than U-Net. Thus, EDPCNN provides a more consistent or stable segmentation. Some segmentation examples predicted by EDPCNN when trained on full dataset are shown in Fig. \ref{fig:example_segmentation} for objects with size from large to small.

Supply of the target object center to EDPCNN can be perceived as a significant advantage. We argue that this advantage cannot overshadow the contribution of end-to-end learning. To establish this claim, we refer readers to Fig. \ref{fig:result_all} and note that the UNet+DP model, despite having the same advantage, lags significantly behind EDPCNN. Therefore, end-to-end learning is the only attributable factor behind the success of EDPCNN.

Further, to test the robustness of EDPCNN with respect to the position of the star pattern center, we perform an experiment where the supplied center during testing is purposely jittered inside the object. Define the radius of an object as the shortest distance from its center to its boundary. We randomly jitter the center supplied to EDPCNN inside a truncated normal distribution that has the boundary defined by the object radius. We do this across five random seeds and plot the mean Dice score across the runs. Fig. \ref{fig:result_robustness} shows the effect of random jitter with the increase of jitter radius from no jitter to 0.5 of the object radius. We can see that there is no significant degradation in performance, especially for 0.2 jitter or below. Fig. \ref{fig:result_robustness} plots the average Dice scores for these experiments. In all the cases, the standard deviation of Dice scores remains small, below 0.01. Thus, the standard deviation has not been shown in Fig. \ref{fig:result_robustness}. 

Finally, Table \ref{tab:time} shows that computationally EDPCNN is about 64\%  more expensive during training than U-Net. However, test time for EDPCNN is only about 16\% more than that of U-Net. 



\section{Summary and Future Work}
In this work, we illustrate how to combine convolutional neural networks and dynamic programming for end-to-end learning. Combination of CNN and traditional tools is not new; however, the novelty here is to handle a \textit{non-differentiable} module, dynamic programming, within the end-to-end pipeline. We employ a neural network to approximate the subgradient of the non-differentiable module. We found that the approximating neural network should have an \textit{exploration} mechanism to be successful.

As a significant application we choose left ventricle segmentation from short axis MRI. Our experiments show that end-to-end combination is beneficial when training data size is small. Our end-to-end model has very little computational overhead, making it a practical choice.

In the future, we plan to segment myocardium and right ventricle with automated placement of star patterns. For these and many other segmentation tasks in medical image analysis, strong object models given by traditional functional modules, such as dynamic programming, provide a way to cope with the lack of training data. Our presented method has the potential to become a blueprint to expand differentiable programming to include \textit{non-differentiable} modules.

\section*{Acknowledgements}
Authors acknowledge funding support from NSERC and Computing Science, University of Alberta.

\bibliographystyle{plainnat}
\bibliography{prletters-template-with-authorship}

\begin{thebibliography}{31}
\providecommand{\natexlab}[1]{#1}
\providecommand{\url}[1]{\texttt{#1}}
\expandafter\ifx\csname urlstyle\endcsname\relax
  \providecommand{\doi}[1]{doi: #1}\else
  \providecommand{\doi}{doi: \begingroup \urlstyle{rm}\Url}\fi

\bibitem[Acton and Ray(2009)]{Acton2009}
Scott~T. Acton and Nilanjan Ray.
\newblock Biomedical image analysis: Segmentation.
\newblock \emph{Synthesis Lectures on Image, Video, and Multimedia Processing},
  4\penalty0 (1):\penalty0 1--108, 2009.
\newblock \doi{10.2200/S00133ED1V01Y200807IVM009}.
\newblock URL \url{https://doi.org/10.2200/S00133ED1V01Y200807IVM009}.

\bibitem[{Bahdanau} et~al.(2014){Bahdanau}, {Cho}, and {Bengio}]{Soft_assn}
D.~{Bahdanau}, K.~{Cho}, and Y.~{Bengio}.
\newblock {Neural Machine Translation by Jointly Learning to Align and
  Translate}.
\newblock \emph{ArXiv e-prints}, September 2014.

\bibitem[Baumgartner et~al.(2018)Baumgartner, Koch, Pollefeys, and
  Konukoglu]{LV_seg}
Christian~F. Baumgartner, Lisa~M. Koch, Marc Pollefeys, and Ender Konukoglu.
\newblock An exploration of 2d and 3d deep learning techniques for cardiac mr
  image segmentation.
\newblock In Mihaela Pop et~al., editors, \emph{Statistical Atlases and
  Computational Models of the Heart. ACDC and MMWHS Challenges}, pages
  111--119, Cham, 2018. Springer Int. Publishing.
\newblock ISBN 978-3-319-75541-0.

\bibitem[Baydin et~al.(2018)Baydin, Pearlmutter, Radul, and Siskind]{Baydin}
Atilim~Gunes Baydin, Barak~A. Pearlmutter, Alexey~Andreyevich Radul, and
  Jeffrey~Mark Siskind.
\newblock Automatic differentiation in machine learning: a survey.
\newblock \emph{Journal of Machine Learning Research}, 18\penalty0
  (153):\penalty0 1--43, 2018.
\newblock URL \url{http://jmlr.org/papers/v18/17-468.html}.

\bibitem[Bernard et~al.(2018)Bernard, Lalande, Zotti, Cervenansky, Yang, Heng,
  Cetin, Lekadir, Camara, Ballester, et~al.]{bernard2018deep}
Olivier Bernard, Alain Lalande, Clement Zotti, Frederick Cervenansky, Xin Yang,
  Pheng-Ann Heng, Irem Cetin, Karim Lekadir, Oscar Camara, Miguel
  Angel~Gonzalez Ballester, et~al.
\newblock Deep learning techniques for automatic mri cardiac multi-structures
  segmentation and diagnosis: Is the problem solved?
\newblock \emph{IEEE Transactions on Medical Imaging}, 2018.

\bibitem[Brosch et~al.(2016)Brosch, Tang, Yoo, Li, Traboulsee, and
  Tam]{Lesion_seg}
T.~Brosch, L.~Y.~W. Tang, Y.~Yoo, D.~K.~B. Li, A.~Traboulsee, and R.~Tam.
\newblock Deep 3d convolutional encoder networks with shortcuts for multiscale
  feature integration applied to multiple sclerosis lesion segmentation.
\newblock \emph{IEEE Transactions on Medical Imaging}, 35\penalty0
  (5):\penalty0 1229--1239, May 2016.
\newblock ISSN 0278-0062.
\newblock \doi{10.1109/TMI.2016.2528821}.

\bibitem[Dou et~al.(2016)Dou, Chen, Yu, Zhao, Qin, Wang, Mok, Shi, and
  Heng]{Deep_detection}
Q.~Dou, H.~Chen, L.~Yu, L.~Zhao, J.~Qin, D.~Wang, V.~C. Mok, L.~Shi, and
  P.~Heng.
\newblock Automatic detection of cerebral microbleeds from mr images via 3d
  convolutional neural networks.
\newblock \emph{IEEE Trans. Med. Im.}, 35\penalty0 (5):\penalty0 1182--1195,
  May 2016.
\newblock ISSN 0278-0062.

\bibitem[Felzenszwalb and Zabih(2011)]{Felzenszwalb}
P.~F. Felzenszwalb and R.~Zabih.
\newblock Dynamic programming and graph algorithms in computer vision.
\newblock \emph{IEEE Transactions on Pattern Analysis and Machine
  Intelligence}, 33\penalty0 (4):\penalty0 721--740, April 2011.
\newblock ISSN 0162-8828.
\newblock \doi{10.1109/TPAMI.2010.135}.

\bibitem[Ghosal and Ray(2017)]{GHOSAL201781}
Sayan Ghosal and Nilanjan Ray.
\newblock Deep deformable registration: Enhancing accuracy by fully
  convolutional neural net.
\newblock \emph{Pattern Recognition Letters}, 94:\penalty0 81 -- 86, 2017.
\newblock ISSN 0167-8655.
\newblock \doi{https://doi.org/10.1016/j.patrec.2017.05.022}.

\bibitem[{Ghosh} et~al.(2017){Ghosh}, {Boulanger}, {Acton}, {Blemker}, and
  {Ray}]{Ghosh2017}
S.~{Ghosh}, P.~{Boulanger}, S.~T. {Acton}, S.~S. {Blemker}, and N.~{Ray}.
\newblock Automated 3d muscle segmentation from mri data using convolutional
  neural network.
\newblock In \emph{2017 IEEE International Conference on Image Processing
  (ICIP)}, pages 4437--4441, Sep. 2017.
\newblock \doi{10.1109/ICIP.2017.8297121}.

\bibitem[Glasmachers(2017)]{Glasmachers17}
Tobias Glasmachers.
\newblock Limits of end-to-end learning.
\newblock \emph{CoRR}, abs/1704.08305, 2017.
\newblock URL \url{http://arxiv.org/abs/1704.08305}.

\bibitem[Greenspan et~al.(2016)Greenspan, van Ginneken, and Summers]{Greenspan}
H.~Greenspan, B.~van Ginneken, and R.~M. Summers.
\newblock Guest editorial deep learning in medical imaging: Overview and future
  promise of an exciting new technique.
\newblock \emph{IEEE Transactions on Medical Imaging}, 35\penalty0
  (5):\penalty0 1153--1159, May 2016.
\newblock ISSN 0278-0062.

\bibitem[He et~al.(2017)He, Mao, Guo, and Yi]{Cell_tracking}
Tao He, Hua Mao, Jixiang Guo, and Zhang Yi.
\newblock Cell tracking using deep neural networks with multi-task learning.
\newblock \emph{Image and Vision Computing}, 60:\penalty0 142 -- 153, 2017.
\newblock ISSN 0262-8856.

\bibitem[Hornik et~al.(1990)Hornik, Stinchcombe, and
  White]{Hornik1990UniversalAO}
Kurt Hornik, Maxwell~B. Stinchcombe, and Halbert White.
\newblock Universal approximation of an unknown mapping and its derivatives
  using multilayer feedforward networks.
\newblock \emph{Neural Networks}, 3:\penalty0 551--560, 1990.

\bibitem[Hu et~al.(2017)Hu, Shuai, Liu, and Wang]{DeepLevelset}
P.~Hu, B.~Shuai, J.~Liu, and G.~Wang.
\newblock Deep level sets for salient object detection.
\newblock In \emph{2017 IEEE Conference on Computer Vision and Pattern
  Recognition (CVPR)}, pages 540--549, July 2017.
\newblock \doi{10.1109/CVPR.2017.65}.

\bibitem[Jaderberg et~al.(2016)Jaderberg, Czarnecki, Osindero, Vinyals, Graves,
  and Kavukcuoglu]{SG2016}
Max Jaderberg, Wojciech~Marian Czarnecki, Simon Osindero, Oriol Vinyals, Alex
  Graves, and Koray Kavukcuoglu.
\newblock Decoupled neural interfaces using synthetic gradients.
\newblock \emph{CoRR}, abs/1608.05343, 2016.
\newblock URL \url{http://arxiv.org/abs/1608.05343}.

\bibitem[Ker et~al.(2018)Ker, Wang, Rao, and Lim]{DL_MIA}
J.~Ker, L.~Wang, J.~Rao, and T.~Lim.
\newblock Deep learning applications in medical image analysis.
\newblock \emph{IEEE Access}, 6:\penalty0 9375--9389, 2018.
\newblock ISSN 2169-3536.
\newblock \doi{10.1109/ACCESS.2017.2788044}.

\bibitem[Kingma and Ba(2014)]{kingma2014adam}
Diederik~P Kingma and Jimmy Ba.
\newblock Adam: A method for stochastic optimization.
\newblock \emph{arXiv preprint arXiv:1412.6980}, 2014.

\bibitem[Le et~al.(2018)Le, Quach, Luu, Duong, and Savvides]{LevelsetRNN}
T.~H.~N. Le, K.~G. Quach, K.~Luu, C.~N. Duong, and M.~Savvides.
\newblock Reformulating level sets as deep recurrent neural network approach to
  semantic segmentation.
\newblock \emph{IEEE Transactions on Image Processing}, 27\penalty0
  (5):\penalty0 2393--2407, May 2018.
\newblock ISSN 1057-7149.

\bibitem[{Marcos} et~al.(2018){Marcos}, {Tuia}, {Kellenberger}, {Zhang}, {Bai},
  {Liao}, and {Urtasun}]{EndToEndAC}
D.~{Marcos}, D.~{Tuia}, B.~{Kellenberger}, L.~{Zhang}, M.~{Bai}, R.~{Liao}, and
  R.~{Urtasun}.
\newblock {Learning deep structured active contours end-to-end}.
\newblock \emph{ArXiv e-prints}, March 2018.

\bibitem[Ngo et~al.(2017)Ngo, Lu, and Carneiro]{NGO2017159}
Tuan~Anh Ngo, Zhi Lu, and Gustavo Carneiro.
\newblock Combining deep learning and level set for the automated segmentation
  of the left ventricle of the heart from cardiac cine magnetic resonance.
\newblock \emph{Medical Image Analysis}, 35:\penalty0 159 -- 171, 2017.
\newblock ISSN 1361-8415.
\newblock \doi{https://doi.org/10.1016/j.media.2016.05.009}.

\bibitem[Pereira et~al.(2016)Pereira, Pinto, Alves, and Silva]{Tumor_seg}
S.~Pereira, A.~Pinto, V.~Alves, and C.~A. Silva.
\newblock Brain tumor segmentation using convolutional neural networks in mri
  images.
\newblock \emph{IEEE Transactions on Medical Imaging}, 35\penalty0
  (5):\penalty0 1240--1251, May 2016.
\newblock ISSN 0278-0062.
\newblock \doi{10.1109/TMI.2016.2538465}.

\bibitem[{Ravishankar} et~al.(2017){Ravishankar}, {Sudhakar}, {Venkataramani},
  {Thiruvenkadam}, {Annangi}, {Babu}, and {Vaidya}]{Ravishankar}
H.~{Ravishankar}, P.~{Sudhakar}, R.~{Venkataramani}, S.~{Thiruvenkadam},
  P.~{Annangi}, N.~{Babu}, and V.~{Vaidya}.
\newblock {Understanding the Mechanisms of Deep Transfer Learning for Medical
  Images}.
\newblock \emph{ArXiv e-prints}, April 2017.

\bibitem[Ray et~al.(2012)Ray, Acton, and Zhang]{DPSnake}
N.~Ray, S.~T. Acton, and H.~Zhang.
\newblock Seeing through clutter: Snake computation with dynamic programming
  for particle segmentation.
\newblock In \emph{Proceedings of the 21st International Conference on Pattern
  Recognition (ICPR2012)}, pages 801--804, Nov 2012.

\bibitem[{Ronneberger} et~al.(2015){Ronneberger}, {Fischer}, and {Brox}]{Unet}
O.~{Ronneberger}, P.~{Fischer}, and T.~{Brox}.
\newblock {U-Net: Convolutional Networks for Biomedical Image Segmentation}.
\newblock \emph{ArXiv e-prints}, May 2015.

\bibitem[Rupprecht et~al.(2016)Rupprecht, Huaroc, Baust, and Navab]{DeepAC}
Christian Rupprecht, Elizabeth Huaroc, Maximilian Baust, and Nassir Navab.
\newblock Deep active contours.
\newblock \emph{CoRR}, abs/1607.05074, 2016.
\newblock URL \url{http://arxiv.org/abs/1607.05074}.

\bibitem[{Shin} et~al.(2016)]{Transfer}
H.-C. {Shin} et~al.
\newblock {Deep Convolutional Neural Networks for Computer-Aided Detection: CNN
  Architectures, Dataset Characteristics and Transfer Learning}.
\newblock \emph{ArXiv e-prints}, February 2016.

\bibitem[Tajbakhsh et~al.(2016)Tajbakhsh, Shin, Gurudu, Hurst, Kendall, Gotway,
  and Liang]{Tajbakhsh}
N.~Tajbakhsh, J.~Y. Shin, S.~R. Gurudu, R.~T. Hurst, C.~B. Kendall, M.~B.
  Gotway, and J.~Liang.
\newblock Convolutional neural networks for medical image analysis: Full
  training or fine tuning?
\newblock \emph{IEEE Trans. on Med. Im.}, 35\penalty0 (5):\penalty0 1299--1312,
  May 2016.
\newblock ISSN 0278-0062.

\bibitem[Tang et~al.(2017)Tang, Valipour, Zhang, Cobzas, and
  J{\"{a}}gersand]{TangVZCJ17}
Min Tang, Sepehr Valipour, Zichen~Vincent Zhang, Dana Cobzas, and Martin
  J{\"{a}}gersand.
\newblock A deep level set method for image segmentation.
\newblock \emph{CoRR}, abs/1705.06260, 2017.
\newblock URL \url{http://arxiv.org/abs/1705.06260}.

\bibitem[V et~al.(2016)V, L, M, and et~al]{retinopathy}
Gulshan V, Peng L, Coram M, and et~al.
\newblock Development and validation of a deep learning algorithm for detection
  of diabetic retinopathy in retinal fundus photographs.
\newblock \emph{JAMA}, 316\penalty0 (22):\penalty0 2402--2410, 2016.
\newblock \doi{10.1001/jama.2016.17216}.

\bibitem[Zotti et~al.(2018)Zotti, Luo, Lalande, and Jodoin]{ShapePriorCNN}
C.~Zotti, Z.~Luo, A.~Lalande, and P.~Jodoin.
\newblock Convolutional neural network with shape prior applied to cardiac mri
  segmentation.
\newblock \emph{IEEE Journal of Biomedical and Health Informatics}, pages 1--1,
  2018.
\newblock ISSN 2168-2194.
\newblock \doi{10.1109/JBHI.2018.2865450}.

\end{thebibliography}

\end{document}